%% file: WebMedia-CTD-CTIC-WFA.tex
\documentclass[sigplan]{webmedia}

\AtBeginDocument{%
  \providecommand\BibTeX{{%
    \normalfont B\kern-0.5em{\scshape i\kern-0.25em b}\kern-0.8em\TeX}}}

\setevent{wfa}  

\acmConference[Anais Estendidos do WebMedia’2023]{}{Ribeirão Preto}{Brasil} 
\ISSN{2596-1683}

\begin{document}

\title{Saturn Platform: Foundation Model Operations and Generative AI for Financial Services}

\author{Antonio J. G. Busson}
\affiliation{%
  \institution{BTG Pactual}
}
\email{antonio.busson@btgpactual.com}

\author{Rennan Gaio}
\affiliation{%
  \institution{BTG Pactual}
}
\email{rennan.gaio@btgpactual.com}

\author{Rafael H. Rocha}
\affiliation{%
  \institution{BTG Pactual}
}
\email{rafael-h.rocha@btgpactual.com}

\author{Francisco Evangelista}
\affiliation{%
  \institution{BTG Pactual}
}
\email{francisco.evangelista@btgpactual.com}

\author{Bruno Rizzi}
\affiliation{%
  \institution{BTG Pactual}
}
\email{bruno.rizzi@btgpactual.com}

\author{Luan Carvalho}
\affiliation{%
  \institution{BTG Pactual}
}
\email{luan.carvalho@btgpactual.com}

\author{Rafael Miceli}
\affiliation{%
  \institution{BTG Pactual}
}
\email{rafael.miceli@btgpactual.com}

\author{Marcos Rabaioli}
\affiliation{%
  \institution{BTG Pactual}
}
\email{marcos.rabaioli@btgpactual.com}

\author{David Favaro}
\affiliation{%
  \institution{BTG Pactual}
}
\email{david.favaro@btgpactual.com}


\renewcommand{\shortauthors}{Busson et al.}




\input{chaps/0_abstract}

\maketitle

\input{chaps/1_introduction}

\input{chaps/2_state_of_the_art}

\input{chaps/3_saturn}
\input{chaps/4_impact}

\input{chaps/5_conclusion}

\bibliographystyle{ACM-Reference-Format}
\bibliography{sample-base}


\end{document}

%% file: chaps/0_abstract.tex
\begin{abstract}
Saturn is an innovative platform that assists Foundation Model (FM) building and its integration with IT operations (Ops). It is custom-made to meet the requirements of data scientists, enabling them to effectively create and implement FMs while enhancing collaboration within their technical domain. By offering a wide range of tools and features, Saturn streamlines and automates different stages of FM development, making it an invaluable asset for data science teams. This white paper introduces prospective applications of generative AI models derived from FMs in the financial sector.
\end{abstract}

%% file: chaps/1_introduction.tex
\section{Introduction}

An emerging Artificial Intelligence (AI) paradigm called the Foundation Model (FM) has shown great potential due to its ability to learn universal representations that can be applied to diverse tasks~\cite{zhou2023comprehensive}. From a technological point of view, foundational models consist of deep learning models that are pre-trained in a self-supervised/semi-supervised manner on a large scale and then adapted for various downstream tasks~\cite{bommasani2021opportunities}.

FMs originated breakthrough innovations in Generative Artificial Intelligence (GenAI) applications like ChatGPT\footnote{https://openai.com/chatgpt} and Bard\footnote{https://bard.google.com/}. GenAI can improve worker productivity by automating repetitive tasks and accelerating content creation. Some studies have found that GenAI could impact 80\% of workers in the United States~\cite{eloundou2023gpts}, increase annual global GDP by 7\% ~\cite{hatzius2023potentially}, and is projected to lead to a 40\% growth in demand for AI and Machine Learning Specialists~\cite{di2023future}.

To maximize the economic advantages of GenAI, businesses must proactively integrate it into their operations. This entails adopting new tools and evolving business models, products, and services to align with this transformative technology~\cite{kanbach2023genai, mariani2022generative}. By doing so, companies can boost the productivity of existing processes and unlock new avenues for growth and innovation.

The development of FMs relies on several significant challenges related to infrastructure, development kits, governance, security, etc. To address these challenges, we propose Saturn, a platform to help the process of building, managing, and serving FMs. Saturn combines advanced technologies, intelligent automation, and robust infrastructure to empower professionals to pursue accurate and reliable models. Saturn's core was generically designed to facilitate its implantation of FMs in any application domain. However, this work will focus on the use cases in the financial sector.

The remainder of this white paper is structured as follows. First, in Section 2, we present gathered requirements for the Saturn platform by analyzing the state-of-the-art in the field of FMs and Generative AI. Next, in Section 3, we describe the Saturn platform architecture. Section 4 explores various use cases of applications within the financial services sector. Finally, Section 5 concludes with our final thoughts and considerations.

The Saturn Platform is proprietary software. All rights are reserved and protected by the BTG Pactual.

%% file: chaps/2_state_of_the_art.tex
\section{Foundation Models and Generative AI}

\subsection{State-of-the-Art}

Foundational models are grounded by two techniques: (1) transfer learning and (2) self-supervised learning. The idea of transfer learning is to apply the knowledge that was learned in training from one task to another different task. On the other hand, in self-supervised learning, the pre-training task is automatically derived from unannotated data. For example, the masked language modeling task used to train BERT~\cite{lee2018pre} is to predict missing words in a sentence.

Self-supervised tasks are more scalable and potentially helpful than models trained in a limited space by label annotation. There has been considerable progress in self-supervised learning since word embedding~\cite{mikolov2013efficient}, which associated each word with a context-independent vector and provided the basis for a wide range of NLP models.

Shortly after, self-supervised methods based on autoregressive language modeling (predicting the next word given the previous words)~\cite{dai2015semi} became popular. This produced models representing contextualized words such as GPT~\cite{radford2018improving}, ELMo~\cite{peters2018deep}, and ULMFiT~\cite{howard2018universal}. The second generation of models based on self-supervised learning, BERT~\cite{kenton2019bert}, GPT-2~\cite{radford2019language}, RoBERTa~\cite{liu2019roberta}, T5~\cite{raffel2020exploring}, and BART~\cite{lewis2019bart} were based on the Transformer architecture, incorporating deeper and more powerful sentence bidirectional encoders.

Inevitably, foundational models underwent a process of homogenization of architectures since the last generation models are all Transformer derivatives ~\cite{vaswani2017attention}. While this homogenization produces hugely high leverage (improvements to the basic models can bring immediate benefits for most of the other foundational models), all AI systems can inherit the same problematic biases from some foundational models~\cite{navigli2023biases, healy2023approaches}.

In addition to the NLP area, methods have been homogenized among research communities in recent years. For example, similar Transformer-based modeling approaches have been applied to images~\cite{dosovitskiy2020image, chen2020generative, oquab2023dinov2}, speech~\cite{liu2020mockingjay}, tabular data~\cite{huang2020tabtransformer, padhi2021tabular}, protein sequences~\cite{rives2021biological}, organic molecules~\cite{rothchild2021c5t5}, and reinforcement learning~\cite{janner2021offline,chen2021decision}. These examples point in a direction where we will have a unified set of tools to develop foundational models for a wide range of modalities~\cite{tamkin2020viewmaker}.

In addition to homogenization, foundational models are a natural way of merging various types of relevant information. Data are naturally multi-modal in some domains, such as foundational models trained on language and vision data ~\cite{liu2022competence, luo2020univl}, tabular and time series~\cite{padhi2021tabular}, tabular and language~\cite{gu2021package}.

Reinforcement Learning has been applied to enhance various FMs in NLP tasks. InstructGPT proposes a method called RLHF (Reinforcement Learning from Human Feedback), which involves fine-tuning large models using PPO (Proximal Policy Optimization) based on a reward model trained to align the models with human preferences~\cite{ouyang2022training}. ChatGPT also employs this approach. The reward model is trained using comparison data generated by human labelers who rank the model outputs manually. Based on these rankings, the reward model or a machine labeler calculates a reward that is then utilized to update the FM through PPO.

A notable advancement in FM technology is GPT-4~\cite{openai2023gpt4}. GPT-4 employs a pre-training phase where it predicts the next token in a document, followed by RLHF fine-tuning. GPT-4 surpasses GPT-3.5 in terms of reliability, creativity, and ability to handle more detailed instructions as the complexity of the task increases.

\subsection{Requirements}

Based on the research in the preceding section, the compiled requirements are as follows:

\begin{enumerate}
    \item Pre-defined self-supervised learning pipelines or frameworks;
    \item Transfer learning support;
    \item Efficient data processing and training pipelines to handle large amounts of unannotated data;
    \item Framework that allows data scientists to build and improve upon existing FMs easily;
    \item Mechanisms to detect and mitigate problematic biases inherited from FMs;
    \item Support a wide range of data modalities, including text, images, speech, tabular data, etc.;
    \item Collaborative features, such as version control, model sharing, and experiment tracking, to facilitate collaboration among data scientists;
    \item Performance optimization by utilizing parallel computing, distributed training, and hardware acceleration (e.g., GPUs or TPUs);
    \item Tools for monitoring model performance in real-world settings and providing insights into potential drift or degradation;
    \item Process for deploying trained FMs into production environments. This includes model-serving infrastructure, REST API support, and containerization for easy integration with other applications;
    \item Support approaches like RLHF for fine-tuning FMs using reward models aligned with human preferences.
\end{enumerate}




%% file: chaps/3_saturn.tex
\section{Saturn Platform}


Saturn platform streamlines the development and deployment of FMs, a critical component for generative AI. Data scientists can efficiently harness their technical expertise with Saturn's comprehensive suite of tools, simplifying the model development process. The platform's automation capabilities further optimize collaboration within data science teams, enhancing the efficiency and effectiveness of LLMs and other generative model applications.

Figure 1 shows the architecture of the Saturn platform. The platform is structured in three environments: (1) Saturn Environment, (2) Data Science (DS) Development,  And (3) Automated FM Operations. Each domain is detailed in the subsections that follow.

\begin{figure*}[t]
    \centering
    \includegraphics[width=0.8\textwidth]{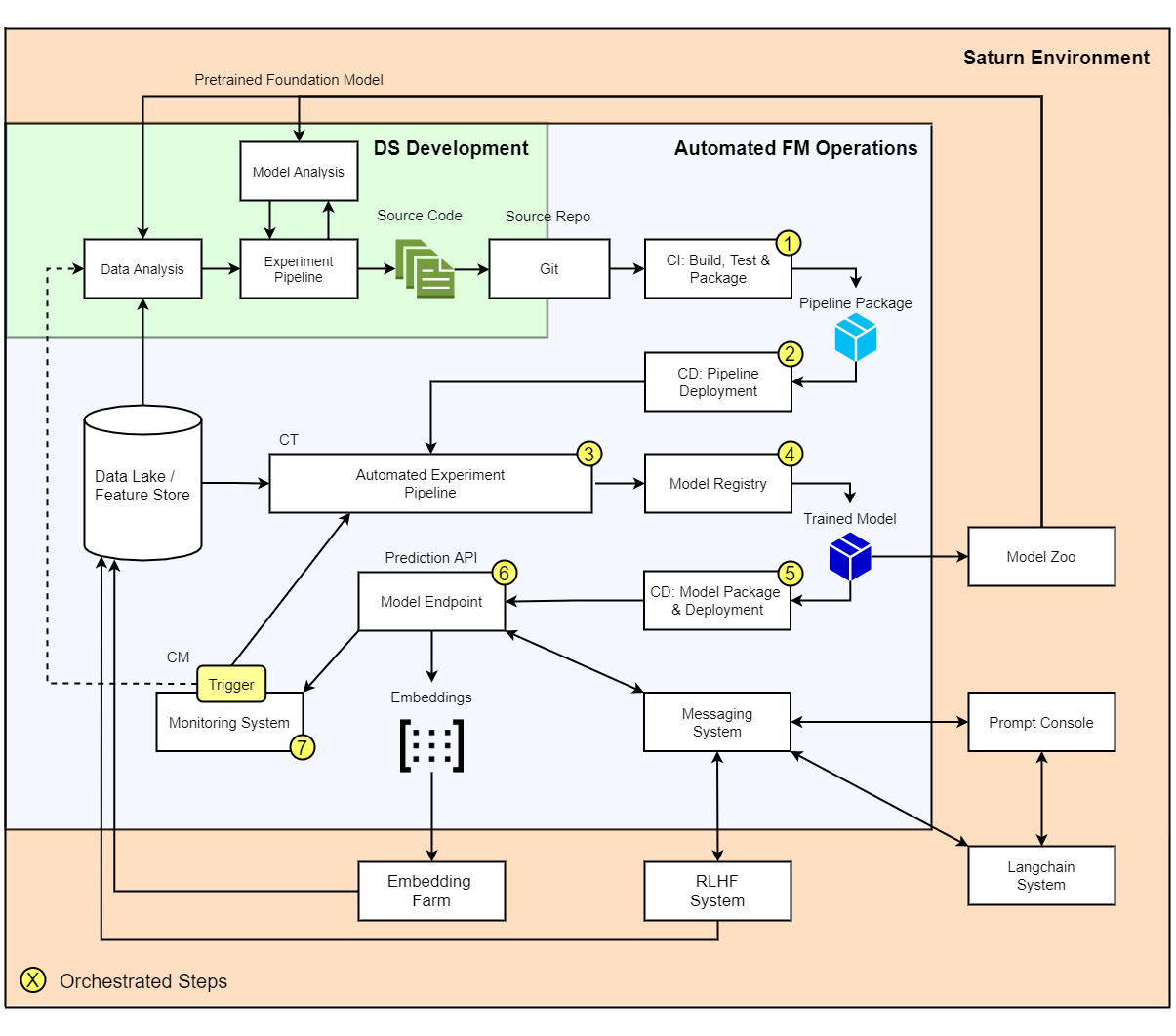}
    \caption{Architecture of the Saturn platform.}
    \label{fig:model_architecture}
\end{figure*}


\subsection{Saturn Environment}

\textit{Model Zoo} offers a centralized repository for FMs, allowing data scientists to access and leverage pre-trained models, architectures, and components. This facilitates knowledge sharing and reduces redundancy, enabling teams to accelerate model development. \textit{Model Zoo} prioritizes data security and governance, ensuring that sensitive data and models are protected throughout the development and deployment life cycle. It includes robust access controls, encryption mechanisms, and compliance features, adhering to industry standards and regulations.

\textit{Embedding Farm} provides an efficient storage solution for embeddings generated by FMs. It leverages advanced algorithms and optimized data structures to ensure high-speed retrieval (via vector database). It offers powerful management capabilities, allowing users to organize, categorize, and tag embeddings. Additionally, \textit{Embedding Farm} incorporates access control mechanisms, ensuring that sensitive embeddings are safeguarded from unauthorized access. 

Saturn can deploy FMs as API endpoints, enabling other software to easily access and leverage these models' power. Whether it is a console prompt or a \textit{LangChain System}\footnote{\url{https://docs.langchain.com/docs/}}, Saturn offers a versatile solution that can be seamlessly integrated into various financial service applications. In addition, the \textit{RLHF system} collects and manages human feedback data to train reward models used to fine-tune FMs via reinforcement learning.

\subsection{DS Development}

The platform provides a collaborative workspace equipped with powerful development tools and libraries tailored for foundation model building. Data scientists can leverage a range of machine learning frameworks and visualization tools, ensuring flexibility and compatibility with their preferred workflows.

In the \textit{Data Analysis} step, pre-trained FMs can assist in the initial data exploration and pre-processing stages. They can generate summaries of large datasets, identify patterns, generate data, and perform basic data-cleaning tasks. This can significantly speed up the data preparation phase, allowing data scientists to focus on higher-level analysis. 

\begin{figure}[t]
    \centering
    \includegraphics[width=0.5\textwidth]{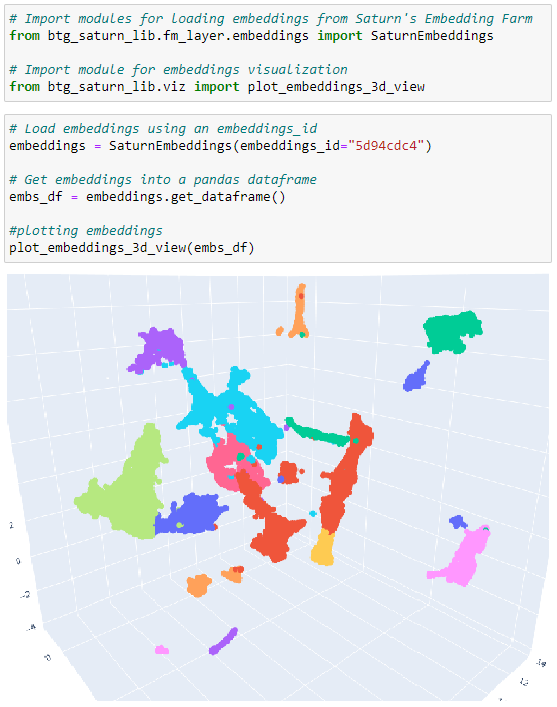}
    \caption{Using SaturnLib to load pre-computed contextual embeddings from Saturn's Embedding Farm.}
    \label{fig:saturn_embeddings}
\end{figure}

The cyclical process between \textit{Experiment Pipeline} definition and \textit{Model Analysis} ensures that data science projects are continuously refined and optimized. Using Saturn, data scientists can iterate quickly on their specific tasks. Instead of training a model from scratch, they can start with a pre-trained FM loaded from \textit{Model Zoo} and fine-tune it, significantly reducing the training time and effort required. This accelerated iteration process facilitates faster experimentation and hypothesis testing.

Loading pre-trained models or pre-computed embeddings on-demand dynamically through Saturno's Python library (SaturnLib) is possible. Users can request specific resources based on keys, and the library fetches them from the database as needed. Figure 2 illustrates an example of using SaturnLib to load pre-computed embeddings. Figure 3 exemplifies an application that loads a pre-trained LLM to create a conversational agent using the Langchain framework.

\begin{figure}[h]
    \centering
    \includegraphics[width=0.5\textwidth]{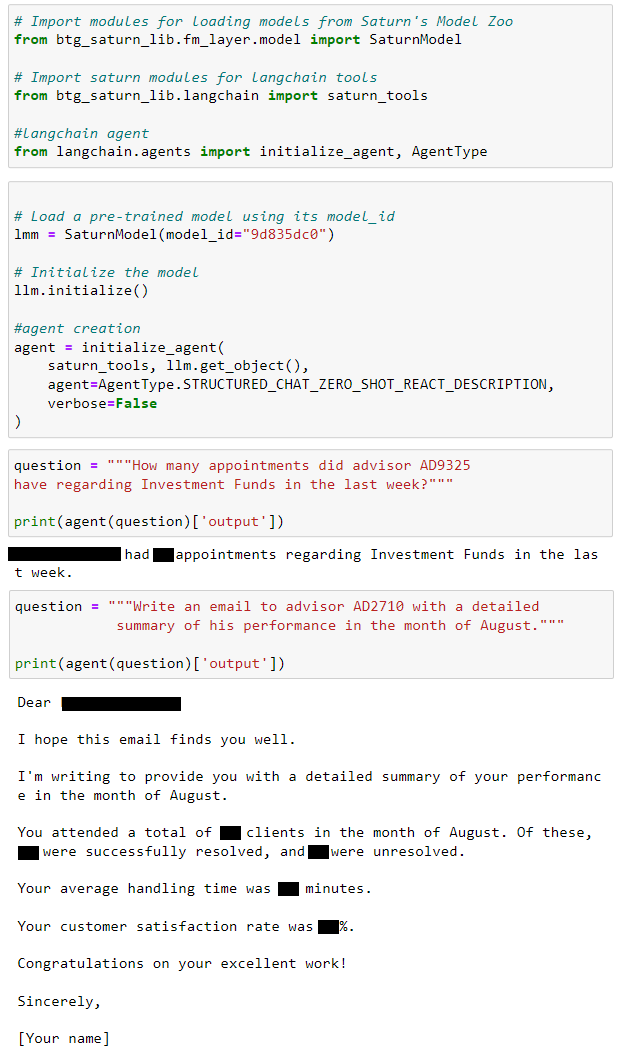}
    \caption{Saturn offers tools compatible with Langchain for building conversational agents. The user can import a pre-trained LLM from Saturn's \textit{Model Zoo} in this example. This LLM will serve as the cognitive engine for the Langchain Agent~\cite{langchainTools}, working in conjunction with Saturn's capabilities to respond to inquiries regarding internal data, including customer service details.}
    \label{fig:model_architecture}
\end{figure}

\subsection{Automated FM Operations (FMOps)}

With integrated IT operations capabilities, FMOps simplifies the deployment and scaling of FMs. It offers automated provisioning and orchestration, enabling data scientists to transition their models from development to production environments easily. 

The FMOps environment is a crucial component of the FM life cycle. It allows FMs to be continuously trained, updated, and deployed, ensuring the effective operation of models in production environments. The various stages of the FM life cycle are outlined in Table 1, and they are an integral part of the well-structured processes depicted in Figure 1. These processes are organized as follows:

\setlength{\tabcolsep}{10pt} 
\renewcommand{\arraystretch}{1.5} 
\begin{table*}[t]
  \centering
  \begin{tabular}{| l | l | p{120mm} |}
   \hline
   \textbf{\#} & \textbf{Stage} & \textbf{Description}  \\ 
   \hline
    S1 & Pre-training & FMs are initially trained on vast amounts of general data in a self-supervised/semi-supervised manner. For example, in the case of language models, they predict the next word in a sentence. Other data types, such as images, tabular data, or time series data, models can be trained using contrastive learning or other self-supervised techniques to capture relevant patterns and features.  \\
    \hline
    S2 & Fine-tuning & After pre-training, the model is fine-tuned for a specific application. This involves training the model on a more specific and narrower dataset related to the intended task. For instance, a language model can be fine-tuned to summarize text or answer questions. In the case of financial FMs, fine-tuning might involve adapting the model for tasks like recommending investment products or assessing churn risk.   \\
    \hline
    S3 & Pre-release testing & Data scientists conduct thorough testing and validation of the model during development. This includes identifying and addressing issues and ensuring the model performs well on the intended task. Testing in the context of FMs can be more structured and intentional due to the increasing complexity and higher stakes involved.  \\
    \hline
    S4 & Release and Distribution & The FMs are released through APIs or other access methods. Data scientists should provide guidelines to users, outlining the appropriate and responsible use of the model. Models may be available through licensing agreements or partnerships with specific organizations, including usage restrictions, acceptable applications, and requirements to use or interact with the model.  \\
    \hline
    S5 & Post-release monitoring  & After the FM is deployed, data scientists monitor its performance and user interactions. This ongoing monitoring helps identify issues, gather user feedback, and make necessary updates to improve the model's performance. Post-release monitoring is crucial for maintaining the reliability and safety of the model in real-world applications.  \\
    \hline
  \end{tabular}
  \caption{Foundation Model life cycle.}
  \label{tab:1}
\end{table*}

\begin{itemize}
    \item (1-3) The user sends the model's source code to a Git server\footnote{\url{https://git-scm.com/}}, where the code will be stored and versioned. A continuous integration and continuous delivery (CI/CD pipeline) treadmill is set up to trigger a continuous training (CT) process whenever there is a new source code. The model's source aims to train a new model or fine-tune a pre-trained model from the \textit{Model Zoo} (S1 and S2).
    
    \item (3-4) After training/fine-tuning, a validation process is undertaken to evaluate the model's performance and safety before it is deployed in practical applications (S3). Subsequently, the resultant model artifact is preserved and cataloged within the Model Zoo, accompanied by the associated model validation metadata.
    
    \item (5-6) The trained model is deployed to a production environment. This can be done through a cloud infrastructure, containers, or model-specific deployment services. The model is configured to be accessed through an API endpoint, allowing external users to request inferences (S4).
    
    \item (7) A continuous monitoring (CM) system is set up to observe the model's performance in production. Data scientists can track relevant metrics, detect anomalies, and receive alerts, allowing them to address issues and optimize model performance (S5) proactively. In addition, the CM system can detect changes in input data (data drift) and automatically starts continuous training to update the model with the latest data.
    
\end{itemize}


%% file: chaps/4_impact.tex
\section{Applications in Financial Services}

\textbf{Forecasting and Predictive Analytics}. Financial institutions often rely on accurate forecasts and predictions for making informed decisions. FMs can be trained on historical financial data. These models can assist in generating forecasts and scenario analysis, helping financial professionals make more informed investment and strategic decisions~\cite{nguyen2023generative}.

\textbf{Financial Report Generation}. FMs/LLMs can generate reports, summaries, and insights based on financial data. They can automatically extract relevant information from financial statements, filings, or market reports and generate concise summaries, reducing the time and effort required for manual analysis. These generated reports can quickly overview key financial metrics, trends, and investment opportunities, facilitating decision-making processes.

\textbf{Risk Assessment}. FMs can assist in risk assessment by analyzing various data sources, including financial statements, market data, credit ratings, and news articles. By processing this information, they can help identify potential risks, such as credit defaults, market volatility, frauds, or company-specific risks~\cite{papapanagiotou2022drift}. This information can support risk management and help financial professionals make informed decisions regarding investment portfolios and risk mitigation strategies.

\textbf{Financial Data Generation}. FMs can assist in generating synthetic financial data that closely resembles real-world data~\cite{chapman2022towards}. This can be beneficial for various purposes, including testing and validating financial models, conducting simulations, or training machine learning algorithms in a controlled environment.

\textbf{Customer Behavior Modeling}. Contextual embeddings generated by FMs~\cite{rehman2023research} help to identify customer segments' distinct financial behaviors and preferences. Among several applications, embeddings can be used to cluster customers into personas based on their financial behavior, allowing for personalized marketing and product recommendations. Additionally, embeddings can be robust features for machine learning models, enhancing their ability to make accurate predictions in financial tasks. 

\textbf{Speech Analysis}. Combining speech transcription models~\cite{nllbteam2022language} with LLMs can be a powerful approach to creating a sophisticated speech analysis system~\cite{min2023exploring}. Especially when it comes to monitoring customer service attendants' actions and performance, ensuring that advisors adhere to regulatory guidelines, discovering client interests, detecting complaints or potential issues raised by customers, and identifying the emotional tone of customer conversations~\cite{parra2022classification}.  

\textbf{Information Retrieval}. LLMs such as Text2SQL models~\cite{kumar2022deep, saeed2023querying, sun2023instruction} are designed to convert natural language queries into structured database queries, making retrieving information from large databases easier. Even employees without a strong background in database querying can use text2SQL to access the information they need, democratizing access to data within an organization.

\textbf{Dashboards Authoring}. Authoring tools using LLMs can provide a natural language interface for creating and customizing dashboards~\cite{srinivasan2023bolt}. Users can interact with the dashboard using plain language commands, making it more accessible to non-technical users. Furthermore, such authoring tools can automate the chart creation process by using data descriptions provided by the user, and the backbone models can generate charts with appropriate labels, colors, and formatting~\cite{maddigan2023chat2vis}.

\textbf{Customer Support and Chatbots}. LLM-powered Chatbots can increase customer engagement and satisfaction by offering personalized content~\cite{hsu2023understanding, zheng2023building, wei2023leveraging}. These chatbots can analyze customer data and previous interactions to personalize responses. This includes tailoring product recommendations, addressing customers by name, and referring to past orders or inquiries. Furthermore, they can be used to answer frequently asked questions and assist customers with routine and technical problems.

\textbf{Legal and Regulatory Compliance}. LLM-based software can analyze vast legal documents, contracts, and regulatory texts~\cite{de2023cicero, huang2023lawyer, trautmann2023large}. They can assist in generating reports by extracting and summarizing relevant information from internal documents, which helps identify potential contract issues and suggest revisions to mitigate risks. Additionally, LLMs can assist in reviewing and generating contracts, ensuring they comply with legal and regulatory standards. 

\textbf{Virtual Assistants}. LLM-powered Agents~\cite{yao2023react, wang2023survey} can be useful in creating virtual assistants for financial advisors. Such intelligent agents can, for instance, draft emails, reports, and other documents in a natural and professional tone, saving advisors time on administrative tasks and improving client communication. Besides, they can provide up-to-date information on market trends, stock prices, economic indicators, and more, helping advisors make informed decisions.

%% file: chaps/5_conclusion.tex
\section{Final Remarks}

Saturn stands out as a pioneering platform leading the charge in the digital transformation of the financial services industry. Its array of functionalities, including pre-trained FMs, contextual embeddings, user-friendly development tools, and resilient infrastructure, provides a decisive edge to data science teams within financial institutions. This advantage is achieved without compromising the stringent security and compliance requirements, underscoring Saturn's significance in this transformative era.

This white paper has introduced a range of prospective applications of generative AI models derived from FMs in the financial sector, emphasizing the potential to revolutionize how financial institutions operate. With Saturn, financial organizations have the opportunity to pioneer innovation and remain at the cutting edge of the industry.